\documentclass[conference,a4paper]{APSIPA2025}
\usepackage{amsmath}
\usepackage{graphicx}
\usepackage{multirow}
\usepackage{booktabs}
\usepackage{wrapfig}
\usepackage{array}
\usepackage{pifont}
\usepackage{amssymb}  
\usepackage[T1]{fontenc}
\usepackage{tabularx,booktabs}
\usepackage[inkscapelatex=false]{svg}
\usepackage{wrapfig}

\usepackage[backend=biber,style=ieee,]{biblatex}
\usepackage{geometry}
\usepackage{fancyhdr}

\fancypagestyle{firststyle}{
  \fancyhf{}
}

\begin{document}

\title{From Blurry to Brilliant Detection: YOLO-Based Aerial Object Detection with Super Resolution}

\author{
\authorblockN{
Ragib Amin Nihal\authorrefmark{1}, 
Benjamin Yen\authorrefmark{1}\authorrefmark{2},
Takeshi Ashizawa\authorrefmark{1},
Katsutoshi Itoyama\authorrefmark{1}, and
Kazuhiro Nakadai\authorrefmark{1}
}
\authorblockA{
\authorrefmark{1}
Systems and Control Engineering, Institute of Science Tokyo, Japan \\
}
\authorblockA{
\authorrefmark{2}
RIKEN, Japan
}
}

\maketitle
\thispagestyle{firststyle}
\pagestyle{fancy}

\begin{abstract}
Aerial object detection presents challenges from small object sizes, high density clustering, and image quality degradation from distance and motion blur. These factors create an information bottleneck where limited pixel representation cannot encode sufficient discriminative features. B2BDet addresses this with a two-stage framework that applies domain-specific super-resolution during inference, followed by detection using an enhanced YOLOv5 architecture. Unlike training-time super-resolution approaches that enhance learned representations, our method recovers visual information from each input image. The approach combines aerial-optimized SRGAN fine-tuning with architectural innovations including an Efficient Attention Module (EAM) and Cross-Layer Feature Pyramid Network (CLFPN). Evaluation across four aerial datasets shows performance gains, with VisDrone achieving 52.5\% mAP using only 27.7M parameters. Ablation studies show that super-resolution preprocessing contributes +2.6\% mAP improvement while architectural enhancements add +2.9\%, yielding +5.5\% total improvement over baseline YOLOv5. The method achieves computational efficiency with 53.8\% parameter reduction compared to recent approaches while achieving strong small object detection performance.
\end{abstract}

\section{Introduction}

Aerial object detection is important for surveillance, agriculture, urban planning,  and  disaster response applications~\cite{nihal2024uav}. Aerial images pose distinct challenges compared to conventional ground-level imagery: objects appear much smaller (often $\le0.3$\% of image area), suffer from motion blur and atmospheric effects, and exist within complex backgrounds~\cite{dota}. 
\\
These challenges affect the visual information available for detection algorithms. When objects occupy only 50-100 pixels or less in aerial imagery, the limited pixel representation cannot encode sufficient discriminative features (detailed shape, texture, and edge information). This creates an \textit{information bottleneck} where the primary limitation is insufficient visual detail in the input imagery, not algorithmic deficiencies.
\\
This information bottleneck manifests as a performance gap between natural and aerial image detection. As shown in Table \ref{tab:performance_gap}, state-of-the-art detectors show reduced accuracy on aerial datasets. For example, YOLOv3 achieves 33.03\% mAP on MS-COCO but only 20.03\% on VisDrone2018 \cite{visdrone}. This drop occurs because natural image datasets like MS-COCO \cite{coco} typically contain large objects against simple backgrounds. Aerial images instead contain smaller, densely clustered objects with greater scale variation and complex environmental conditions.
\\
Recent approaches have addressed aerial detection challenges through specialized architectures \cite{sca,msa,tph}, attention mechanisms \cite{tph++}, and multi-scale feature fusion \cite{dssd,assf}. These methods focus on architectural improvements without addressing the underlying issue: insufficient visual information in small, blurred objects. These methods can improve feature extraction but cannot recover detail absent from the input image.
\\
Super-resolution (SR) techniques can address this information bottleneck. Zhang et al. \cite{SuperYOLO} developed SR-assisted detection through \textit{training-time} super-resolution guidance, where an SR branch guides backbone learning during training but is discarded during inference for computational efficiency. Their approach achieves 73.61\% accuracy on VEDAI dataset \cite{razakarivony2016vehicle} with 18.1× fewer GFLOPs than YOLOv5x by enhancing learned feature representations without inference overhead. However, training-time SR approaches have a limitation: they enhance learned representations but cannot recover visual information absent in low-resolution test images.
\\
\begin{table}[tb]
\centering
\caption{Performance disparity between natural and aerial datasets. Some statistics have been sourced from \cite{Zoominnet}.
}
\resizebox{\columnwidth}{!}{%
\begin{tabular}{|l|c|c|c|c|c|}
\hline
Dataset        & SSD   & YOLOv3 & RefineDet & Faster-RCNN & RetinaDet \\
\hline
MS-COCO  \cite{coco}      & 26.81  & 33.03   & 41.79      & 41.48        & 39.13      \\
VisDrone2018 \cite{visdrone}  & 2.52  & 20.03  & 21.07     & 21.34       & 31.88     \\
\hline
\end{tabular}
}
\label{tab:performance_gap}
\vspace*{-0.9cm}
\end{table}
This raises the \textbf{research question}: \textit{Can domain-specific super-resolution preprocessing combined with architectural adaptations overcome the information bottleneck in aerial object detection? Furthermore, can inference-time processing achieve this more effectively than training-time guidance?} We hypothesize that enhancing visual information through aerial-optimized super-resolution before detection can improve performance compared to architectural improvements alone. This approach addresses the information bottleneck by recovering visual detail that is absent in low-resolution inputs.
\\
To test this hypothesis, we present B2BDet (Blurry to Brilliant Detection), which uses \textit{inference-time super-resolution preprocessing} combined with domain-specific optimization. Unlike training-time SR guidance that enhances learned representations, our approach recovers visual information in each input image during detection.
Our key contributions that distinguish B2BDet from existing SR-assisted detection methods are:
\begin{enumerate}
\item \textbf{Inference-time SR preprocessing}: Unlike training-time SR guidance \cite{SuperYOLO}, we apply super-resolution to each input image during detection, enhancing visual information rather than learned representations. 
\item \textbf{Domain-specific SRGAN optimization}: We develop an aerial-optimized SRGAN fine-tuned on aerial imagery with edge-preserving loss functions specifically designed for small object detection. This addresses the limitation of generic SR models that fail to preserve spatial details for aerial imagery.
\item \textbf{Post-SR architectural enhancements}: We introduce EAM and CLFPN modules designed to leverage information from super-resolved inputs, providing improvements to the preprocessing enhancement.
\item  \textbf{Comprehensive evaluation} across four diverse aerial datasets with systematic ablation studies show that inference-time super-resolution contributes +2.6\% mAP improvement, while architectural enhancements provide additional +2.9\% mAP improvement, total system improvement of +5.5\% mAP over baseline methods.
\end{enumerate}

\section{Related Work}

\subsection{Super-Resolution Assisted Object Detection}

Super-resolution and object detection combinations have been explored to improve small object detection performance. Zhang et al. \cite{SuperYOLO} proposed SuperYOLO, where the approach uses an SR branch to guide feature learning during training, then discards it during inference to avoid computational overhead.
However, training-time approaches enhance learned representations but cannot recover visual information absent in low-resolution test images. The information bottleneck persists at inference time since no actual visual enhancement occurs during detection.
\\
Our approach applies super-resolution directly to input images during inference, recovering visual detail rather than relying solely on enhanced training representations. We also fine-tune the SR model specifically for aerial imagery characteristics, whereas existing approaches use more general remote sensing optimization. Table \ref{tab:approach_comparison} summarizes the key methodological differences between our approach and  related works.
\\

\begin{figure}[t]
\centering
\includegraphics[width=.95\columnwidth]{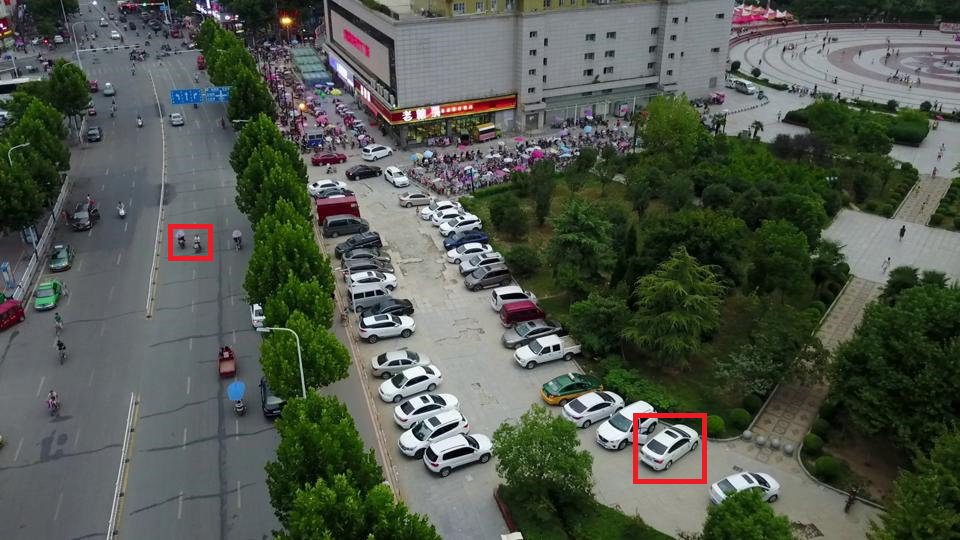}\\
(a)\\[1ex]
\includegraphics[width=.95\columnwidth]{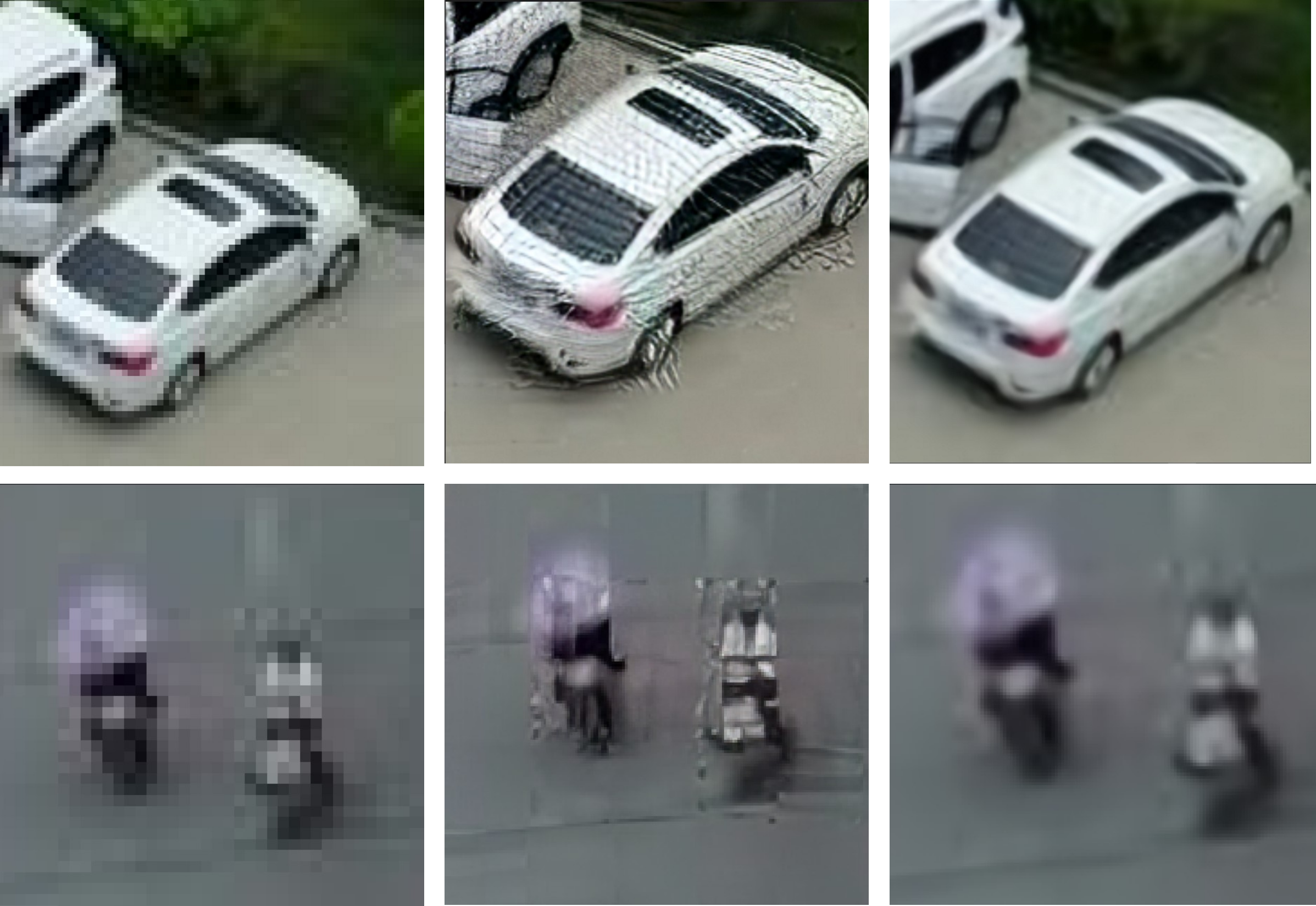}\\
(b)
\caption{Sample aerial image with highlighted regions for super-resolution comparison. (b) Detailed comparison showing original crop (left), pretrained ESRGAN \cite{ESRGAN} result with artifacts and blurring (middle), and our fine-tuned model producing sharper results with better edge preservation and detail recovery optimized for aerial imagery (right).}
\label{sp_comp}
\vspace*{-0.6cm}
\end{figure}
\vspace*{-0.6cm}
\subsection{Aerial Object Detection Architectures}

\noindent Aerial object detection presents challenges including small object sizes, high density clustering, and complex backgrounds \cite{od,od2}. Researchers developed architectures to address these issues. DSSD \cite{dssd} introduces deconvolution layers for enhanced feature resolution, while MSA-YOLO \cite{msa} and SCA-YOLO \cite{sca} employ attention mechanisms to focus on relevant features, with SCA-YOLO achieving 47.4\% mAP on VisDrone.
\\
Transformer-based approaches like TPH-YOLOv5 \cite{tph} integrate transformer prediction heads for global context modeling, achieving 48.9\% mAP on VisDrone. Multi-scale feature fusion techniques combine features from different scales to handle the scale variation common in aerial imagery.
\\
These architectural improvements enhance feature processing but do not address the underlying issue where small objects lack sufficient pixel detail. Our EAM and CLFPN modules complement these approaches by leveraging enhanced visual information available after super-resolution preprocessing.
\vspace*{-0.2cm}
\subsection{Super-Resolution for Aerial Imagery}
Traditional super-resolution methods for natural images often fail on aerial imagery due to different challenges: scale variation, complex textures, and overhead perspective patterns. SRGAN \cite{SRGAN} and ESRGAN \cite{ESRGAN} have shown promise for high-quality image upscaling, but generic pretrained models fail to preserve high-frequency details needed for aerial object detection.
\\
Generic SR models optimize for perceptual quality in natural images rather than preserving fine-grained spatial details needed for small object boundaries in aerial imagery. Domain-specific optimization should instead prioritize edge preservation and spatial detail recovery over general perceptual quality.
\\
Our approach uses aerial-specific SRGAN fine-tuning with edge-preserving loss functions designed to enhance small object detectability rather than general image quality. Figure \ref{sp_comp} illustrates the quality of our domain-specific approach compared to generic pretrained models on aerial imagery.
\begin{table}[b]
\centering
\caption{Methodological comparison with related works}
\label{tab:approach_comparison}
\resizebox{\columnwidth}{!}{%
\begin{tabular}{|l|c|c|c|c|c|}
\hline
\textbf{Characteristic} & \textbf{SuperYOLO} & \textbf{TPH-YOLOv5} & \textbf{SCA-YOLO} & \textbf{DSSD} & \textbf{B2BDet} \\
\hline
SR Enhancement & Training & \ding{55} & \ding{55} & \ding{55} & Inference \\
Attention Mechanism & \ding{55} & Transformer & Spatial & \ding{55} & Efficient \\
Multi-scale Fusion & Standard & Enhanced & Standard & Deconv & Adaptive \\
Input Enhancement & \ding{55} & \ding{55} & \ding{55} & \ding{55} & \checkmark \\
Domain-specific SR & \ding{55} & \ding{55} & \ding{55} & \ding{55} & \checkmark \\
\hline
\end{tabular}
}
\end{table}
\section{Methodology}

\subsection{Problem Formulation}

Let $\mathcal{I} = \{I_1, I_2, ..., I_N\}$ denote aerial images where each image $I_i \in \mathbb{R}^{H \times W \times 3}$ contains objects $\mathcal{O}_i = \{o_1, o_2, ..., o_{K_i}\}$. Each object $o_k$ is characterized by its bounding box coordinates $(x_k, y_k, w_k, h_k)$ and class label $c_k$.
\\
The aerial object detection problem has a key constraint: object area ratio $\rho_k = \frac{w_k \cdot h_k}{H \cdot W} < \tau$ where $\tau = 0.003$ for the majority of instances in aerial datasets. This small object constraint creates an information bottleneck. 
\\
The \textit{information bottleneck} in aerial object detection arises when objects contain insufficient pixels to encode discriminative visual features. For an object with area $A_{obj}$ pixels, the theoretical information capacity is bounded by $\log_2(A_{obj})$ bits. When $A_{obj} < 100$ pixels, which is common in aerial imagery, this capacity becomes limited.
\\
The challenge of object detection can be formally characterized using the concept of mutual information $\mathcal{I}(X; Y)$ between visual features $X$ extracted from the object region and object labels $Y$:
$\mathcal{I}(X; Y) = \mathcal{H}(Y) - \mathcal{H}(Y|X)$
\\
For small aerial objects, the conditional entropy $\mathcal{H}(Y|X)$ remains high because the limited visual features $X$ provide insufficient information to distinguish between object classes. When objects are heavily downsampled, such as a car reduced to 8×8 pixels, the visual features become nearly uninformative:
$\mathcal{H}(Y|X) \approx \mathcal{H}(Y) = \log_2|\mathcal{O}|$
\\
This yields $\mathcal{I}(X; Y) \approx 0$, indicating minimal discriminative information. The information bottleneck comes from the basic limitation that small objects cannot encode sufficient visual detail. Our approach addresses this: super-resolution preprocessing increases the information capacity by enhancing visual detail before detection.
\begin{figure}[t]
\centering
\includegraphics[width=\columnwidth]{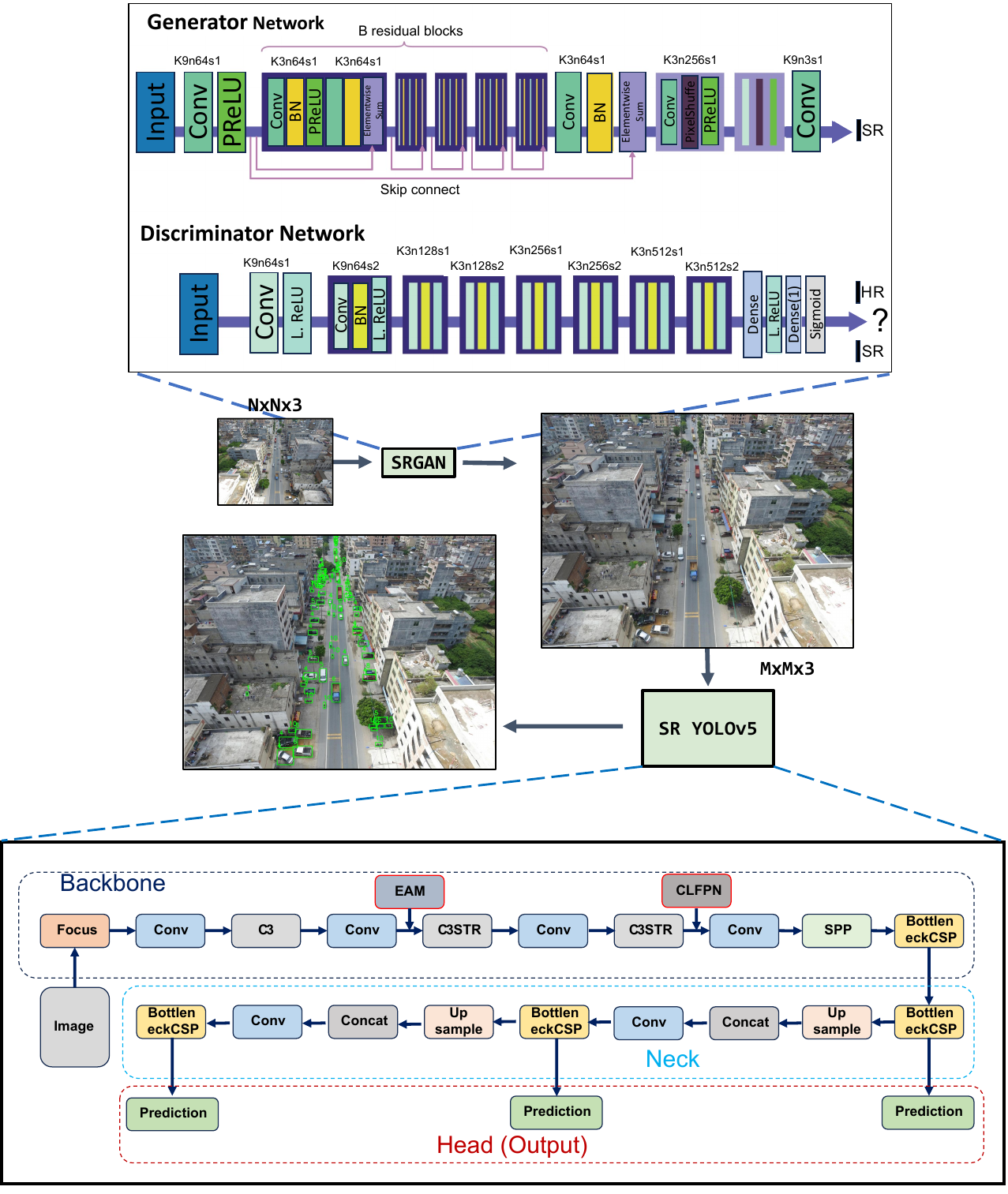}
\caption{B2BDet framework overview showing the two-stage approach: Stage 1 applies aerial-optimized super-resolution using custom SRGAN, Stage 2 performs object detection using enhanced YOLOv5 with EAM and CLFPN modules for improved multi-scale feature processing.}
\label{fig:architecture}
\vspace*{-.50cm}
\end{figure}
\subsection{Aerial-Optimized Super-Resolution Framework}

We formulate aerial super-resolution as a constrained optimization problem. Given low-resolution aerial image $I_{LR} \in \mathbb{R}^{h \times w \times 3}$, we seek to recover high-resolution image $I_{SR} \in \mathbb{R}^{rh \times rw \times 3}$ where $r$ is the scaling factor.
\\
The generator network $G_\theta: \mathbb{R}^{h \times w \times 3} \rightarrow \mathbb{R}^{rh \times rw \times 3}$ is optimized via:
$$\theta^* = \arg\min_\theta \mathbb{E}_{I_{LR} \sim p_{aerial}} \left[ \mathcal{L}_{total}(G_\theta(I_{LR}), I_{HR}) \right]$$
where $p_{aerial}$ represents the distribution of aerial imagery,  assumed distinct from natural image distributions.
\\
The total loss combines pixel fidelity, perceptual quality, and edge preservation:
$$\mathcal{L}_{total} = \mathcal{L}_{MSE} + \lambda_{adv}\mathcal{L}_{adv} + \lambda_{edge}\mathcal{L}_{edge} + \lambda_{perceptual}\mathcal{L}_{perceptual}$$
where the edge loss term  $\mathcal{L}_{edge} = \frac{1}{N}\sum_{i=1}^{N} ||\nabla I_{HR}^{(i)} - \nabla G_\theta(I_{LR}^{(i)})||_1$, is crucial for aerial imagery as edge information carries high-frequency details essential for object boundaries and shapes. For small objects, the relationship between edge clarity and detectability can be expressed as:
$\text{Detectability} \propto \frac{\text{Edge\_Sharpness} \times \text{Perimeter}}{\text{Area}}$
\\
Small objects with well-preserved edges provide higher information density than blurred objects of the same size. Thus, preserving edge gradients through super-resolution directly addresses the information bottleneck by recovering high-frequency spatial details lost in low-resolution imagery.
\\
Fig. \ref{sp_comp} shows the results of our domain-specific fine-tuning approach. Pretrained ESRGAN models exhibit artifacts and blurring when applied to aerial imagery. Our fine-tuned model produces sharper, clearer results with better preservation of edge information and fine details for aerial imagery.
\vspace*{-.3cm}
\subsection{Enhanced YOLOv5 Architecture}
Our custom YOLOv5 architecture addresses specific challenges in aerial images through two key architectural enhancements designed to improve feature representation and multi-scale object detection.
\\
\textbf{Efficient Attention Module (EAM):} We reformulate traditional attention mechanisms for aerial imagery constraints. Given feature map $\mathbf{F} \in \mathbb{R}^{H' \times W' \times D}$, our EAM computes:
$$\text{EAM}(\mathbf{F}) = \mathbf{F} + \sigma\left(\mathbf{W}_v \cdot \text{Attention}(\mathbf{F})\right),$$
$$\text{Attention}(\mathbf{F}) = \text{Softmax}\left(\frac{(\mathbf{W}_q\mathbf{F})(\mathbf{W}_k\mathbf{F})^T}{\sqrt{d_k}}\right)(\mathbf{W}_v\mathbf{F}),$$
where $\mathbf{W}_q, \mathbf{W}_k, \mathbf{W}_v \in \mathbb{R}^{D \times d}$ are learned projection matrices with $d = D/4$ for computational efficiency, achieving 4× reduction compared to standard attention while maintaining representational capacity.
\\
\textbf{Cross-Layer Feature Pyramid Network (CLFPN):} Traditional FPN assumes linear feature combination, insufficient for aerial imagery's extreme scale variation. Our CLFPN implements adaptive cross-scale fusion:
\begin{align*}
\mathbf{Y}_i &= \mathbf{W}_{\mathcal{F}_i} \left( \bigoplus_{j=i}^{L} \alpha_{i,j} \cdot U_{i \leftarrow j}(\mathcal{L}_j) \right), \\
\alpha_{i,j} &= \frac{\exp(\mathbf{w}_{i,j}^T \cdot \text{GAP}(\mathcal{L}_j))}{\sum_{k=i}^{L} \exp(\mathbf{w}_{i,k}^T \cdot \text{GAP}(\mathcal{L}_k))}
\end{align*}

where $\bigoplus$ denotes element-wise addition, $U_{i \leftarrow j}$ represents upsampling from layer $j$ to resolution of layer $i$, $\alpha_{i,j}$ are learned attention weights, and GAP is Global Average Pooling.
\\
For objects with scale $s_k$, the optimal feature resolution $r^*$ minimizes the detection error:
$$r^* = \arg\min_r \mathbb{E}_{s_k}\left[\mathcal{L}_{detection}(s_k, r)\right]$$
Our CLFPN approximates this by adaptively weighting features from multiple resolutions based on content analysis.
Fig. \ref{fig:architecture} illustrates the complete B2BDet framework, showing the integration of our aerial-optimized super-resolution with the enhanced YOLOv5 architecture.
\begin{table}[t]
\centering
\caption{Performance and efficiency comparison on VisDrone dataset}
\label{tab:comparison}
\resizebox{\columnwidth}{!}{%
\begin{tabular}{|l|l|c|c|c|c|c|}
\hline
\textbf{Method} & \textbf{Venue} & \textbf{mAP50} & \textbf{AP$_S$} & \textbf{Params} & \textbf{FLOPs} & \textbf{FPS} \\
& & \textbf{(\%)} & \textbf{(\%)} & \textbf{(M)} & \textbf{(B)} & \\
\hline
Cascaded Zoom-in \cite{meethal2023} & CVPR'23 & 58.3 & 26.1 & - & - & 8.4 \\
TPH-YOLOv5 \cite{tph} & ICCV'21 & 48.9 & 33.8 & 60.0 & 207.0 & 18.5 \\
SCA-YOLO \cite{sca} & VC'23 & 47.4 & 32.1 & 65.1 & 142.3 & 24.2 \\
MSA-YOLO \cite{msa} & Sensors'23 & 34.7 & 21.2 & 58.3 & - & 25.6 \\
YOLOv4 \cite{yolov4} & arXiv'20 & 30.7 & 18.5 & 64.2 & - & 22.8 \\
\hline
\textbf{B2BDet (Ours)} & & \textbf{52.5} & \textbf{38.9} & \textbf{27.7} & \textbf{109.5} & \textbf{27.3} \\
\hline
\end{tabular}
}
\vspace*{-.7cm}
\end{table}
\subsection{Training Protocol}

The components are trained separately. Stage 1 trains SRGAN using alternating gradient descent:
\begin{align*}
\theta_{t+1} &= \theta_t - \eta_{\mathcal{G}} \nabla_\theta \mathcal{L}_{total}(\theta_t, \phi_t), \\
\phi_{t+1} &= \phi_t - \eta_{D} \nabla_\phi \mathcal{L}_{adv}(\theta_t, \phi_t)
\end{align*}
Stage 2 data augmentation techniques include mosaic augmentation with probability 1.0, mixup with probability 0.2, random affine transformations, and color space augmentations.

\section{Experimental Evaluation}

\subsection{Datasets and Implementation}
We evaluate B2BDet on four aerial datasets:
\\
\textbf{VisDrone:} 6,471 drone-captured images with 343,205 object instances, emphasizing small object detection with majority of objects smaller than 0.3\% of image area \cite{visdrone}.\\
\textbf{SeaDroneSee:} 2,900+ maritime images annotated with search and rescue objects like swimmers, boats, and buoys \cite{seadronesee}.
\\
\textbf{VEDAI:} 1,433 images focusing on vehicle detection with annotations for 8 vehicle classes \cite{vedai}.
\\
\textbf{NWPU VHR-10:} 800 geospatial images with 10 classes at very high resolutions of 0.5-2m per pixel \cite{nwpu}.
\\
The super-resolution model was trained on NVIDIA A100 GPU with batch size 16 and 4000 iterations using ADAM with learning rate 1e$^{-4}$. SR-YOLOv5 was trained for 100 epochs. We used cosine scheduling and early stopping.
\begin{table}[t]
\centering
\caption{Comprehensive ablation study on VisDrone dataset}
\label{tab:ablation}
\resizebox{\columnwidth}{!}{%
\begin{tabular}{|l|c|c|c|c|c|}
\hline
\textbf{Method} & \textbf{SR} & \textbf{EAM} & \textbf{CLFPN} & \textbf{mAP50} & \textbf{AP$_S$} \\
& & & & \textbf{(\%)} & \textbf{(\%)} \\
\hline
Baseline YOLOv5 & $\times$ & $\times$ & $\times$ & 47.0 & 28.3 \\
+ Generic SRGAN & $\checkmark$ & $\times$ & $\times$ & 48.1 & 32.1 \\
+ Fine-tuned SRGAN & $\checkmark$ & $\times$ & $\times$ & 49.6 & 35.8 \\
YOLOv5 + EAM & $\times$ & $\checkmark$ & $\times$ & 48.2 & 30.1 \\
YOLOv5 + CLFPN & $\times$ & $\times$ & $\checkmark$ & 48.7 & 31.5 \\
YOLOv5 + EAM + CLFPN & $\times$ & $\checkmark$ & $\checkmark$ & 49.1 & 32.2 \\
\textbf{B2BDet (Full)} & $\checkmark$ & $\checkmark$ & $\checkmark$ & \textbf{52.5} & \textbf{38.9} \\
\hline
\multicolumn{6}{|l|}{\textbf{Component Analysis:}} \\
Super-Resolution Contribution & \multicolumn{4}{|l|}{} & +2.6\% \\
EAM Contribution & \multicolumn{4}{|l|}{} & +1.2\% \\
CLFPN Contribution & \multicolumn{4}{|l|}{} & +1.7\% \\
\textbf{Total System Improvement} & \multicolumn{4}{|l|}{} & \textbf{+5.5\%} \\
\hline
\end{tabular}
}
\vspace*{-.7cm}
\end{table}
\subsection{Evaluation Metrics}

We evaluate our method using standard object detection metrics following the COCO evaluation protocol \cite{coco}. The primary metric is mean Average Precision (mAP), calculated as:
$\text{mAP} = \frac{1}{|C|} \sum_{c=1}^{|C|} \text{AP}_c$,
where $|C|$ is the number of classes and $\text{AP}_c$ is the Average Precision for class $c$, computed as:
$\text{AP}_c = \int_0^1 P_c(R) \, dR$,
where $P_c(R)$ is the precision-recall curve for class $c$. Precision and recall are defined as:
$$\text{Precision} = \frac{\text{TP}}{\text{TP} + \text{FP}}, \quad \text{Recall} = \frac{\text{TP}}{\text{TP} + \text{FN}}$$
where TP, FP, and FN denote true positives, false positives, and false negatives, respectively. A detection is considered a true positive if the Intersection over Union (IoU) with ground truth exceeds a threshold $\tau$:
$$\text{IoU} = \frac{\text{Area}(B_{\text{pred}} \cap B_{\text{gt}})}{\text{Area}(B_{\text{pred}} \cup B_{\text{gt}})}$$
We report mAP@0.5 with IoU threshold $\tau = 0.5$ as the primary metric, along with $\text{AP}_S$ for small objects with area $< 32^2$ pixels. Given the prevalence of small objects in aerial imagery, $\text{AP}_S$ serves as an important indicator of method effectiveness for aerial applications.
\vspace*{-.2cm}
\subsection{Comparison with State-of-the-Art}
Table \ref{tab:comparison} compares B2BDet with recent methods on performance and efficiency. Figure \ref{fig:unlabeled} shows detection in a sample image.
Our method does not achieve the current state-of-the-art of 58.3\% by Cascaded Zoom-in. However, B2BDet achieves good efficiency with 53.8\% fewer parameters than comparable methods. It also achieves the best small object detection performance with $\text{AP}_S$ of 38.9\%. 
\vspace*{-.2cm}
\subsection{Ablation Study and Analysis}

We evaluate each component's contribution through ablation studies on VisDrone. Table \ref{tab:ablation} shows the impact of super-resolution preprocessing and architectural enhancements.
\\
Individual module testing shows EAM provides +1.2\% improvement with stronger benefits for small objects (+1.8\% in $\text{AP}_S$). CLFPN contributes +1.7\% improvement with more balanced improvements across object sizes. Super-resolution preprocessing alone achieves +2.6\% improvement. This improvement is largest for small objects (+7.5\% in $\text{AP}_S$). The complete B2BDet framework achieves +5.5\% improvement over baseline.
\vspace*{-.2cm}
\subsection{Cross-Dataset Evaluation}
We evaluate B2BDet across all four aerial datasets in Table \ref{tab:cross_dataset}. The improvements range from +5.5\% to +8.2\% across different imaging conditions, showing that our approach works well across datasets. The method shows strong performance on maritime scenarios with SeaDroneSee achieving +8.2\% improvement. It also performs well on high-resolution imagery with NWPU VHR-10 achieving +6.8\% improvement. This indicates that the domain-specific super-resolution handles various aerial imaging challenges.
\begin{table}[t]
\centering
\caption{Cross-dataset performance comparison}
\label{tab:cross_dataset}
\scriptsize
\begin{tabular}{|l|c|c|c|}
\hline
\textbf{Dataset} & \textbf{Baseline YOLOv5} & \textbf{B2BDet} & \textbf{Improvement} \\
& \textbf{mAP50 (\%)} & \textbf{mAP50 (\%)} & \textbf{(\%)} \\
\hline
VisDrone & 47.0 & 52.5 & +5.5 \\
SeaDroneSee & 67.8 & 76.0 & +8.2 \\
VEDAI & 71.2 & 77.5 & +6.3 \\
NWPU VHR-10 & 83.7 & 90.5 & +6.8 \\
\hline
\end{tabular}
\vspace*{-.7cm}
\end{table}
\vspace*{-.2cm}
\subsection{Efficiency Analysis}
Table~\ref{tab:comparison} shows that architectural optimizations achieve computational efficiency with large parameter reduction.

\section{Discussion}
\begin{figure}[t]
    \centering
    \includegraphics[width=0.9\columnwidth]{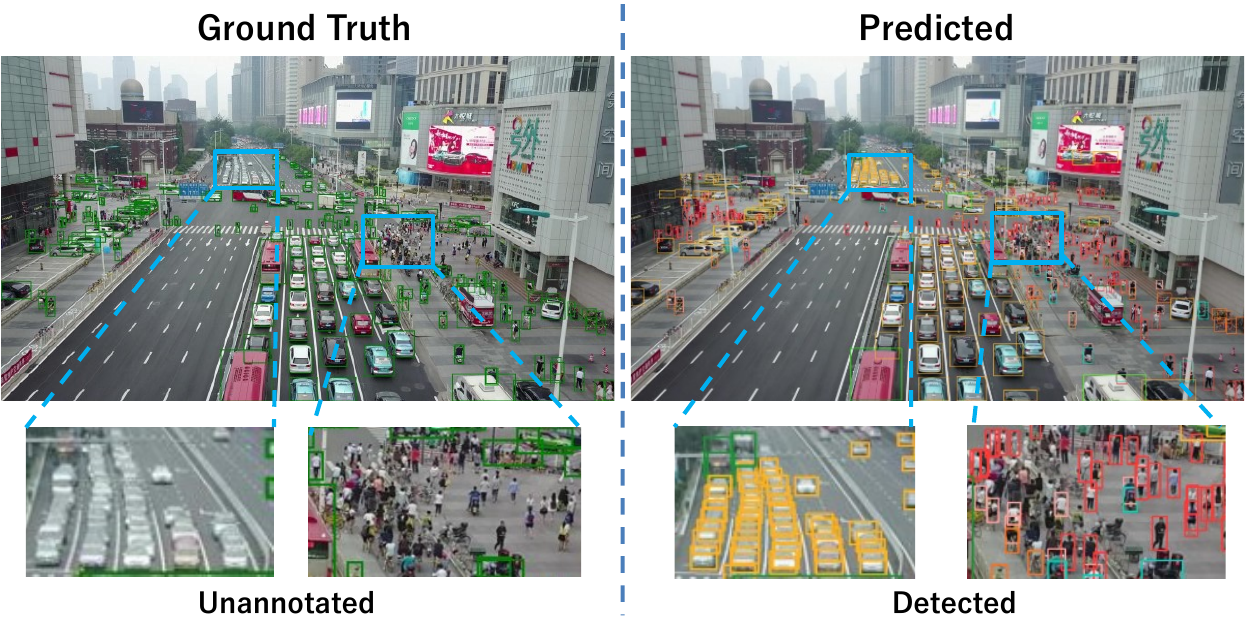}
    \caption{Example from VisDrone dataset where ground truth annotations (left) fail to label certain objects, while our B2BDet (right) successfully detects them, leading to underestimation of actual performance in standard evaluation metrics.}
    \label{fig:unlabeled}
    \vspace*{-.5cm}
\end{figure}

Our approach demonstrates the effectiveness of combining domain-specific super-resolution with architectural optimizations for aerial object detection. The +5.5\% mAP improvement over baseline YOLOv5 shows that preprocessing enhancement and architectural innovation can work combinedly.
\\
However, our method does not achieve the current state-of-the-art performance on VisDrone (58.3\% by Cascaded Zoom-in Detector). Our contribution lies in providing an efficient alternative with significantly fewer parameters (27.7M vs 60M+) while maintaining competitive accuracy. The approach offers a favorable trade-off between accuracy and computational efficiency, important for real-time aerial applications.
\\
\textbf{Key Findings:} The +2.6\% mAP improvement from super-resolution preprocessing validates domain-specific fine-tuning over generic SR models. EAM and CLFPN architectural enhancements complement super-resolution effectively, providing +2.9\% additional improvement. Improvements on challenging small object classes confirm the method's effectiveness for difficult aerial detection scenarios.
\\
\textbf{Training Data Distribution Impact}: We observe a notable correlation between the number of training instances per object class and detection performance. On VisDrone, the car category with over 200,000 training instances achieves high performance (87.8\% AP), while the awning tricycle category with only 4,000 instances results in significantly lower performance (31.7\% AP). This relationship follows an approximate logarithmic trend:
$$\text{AP}_{\text{class}} = \alpha \log(N_{\text{instances}}) + \beta$$
where empirical analysis yields $\alpha \approx 8.3$ and $\beta \approx -12.7$ across VisDrone classes. The data imbalance is challenging for aerial detection where certain object types (vehicles) naturally appear more frequently than others (specialized equipment). Our method shows improved robustness to this imbalance compared to baseline approaches, with smaller performance gaps between well-represented and under-represented classes, though the fundamental data distribution challenge remains.
\\
\textbf{Evaluation Challenges in Aerial Detection:} An interesting observation during our evaluation concerns incomplete annotations in aerial datasets, where our model detects valid objects that are not marked in the ground truth, as illustrated in Fig. \ref{fig:unlabeled}. This phenomenon is observed for small or distant objects like vehicles and pedestrians, where manual annotation can be challenging due to their size and visual similarity to background elements. These detections are counted as false positives in standard evaluation metrics.
This suggests that quantitative improvements may be underestimated and highlights the need for more comprehensive annotation protocols in aerial detection benchmarks.
\\
\noindent\textbf{Limitations:} Performance degrades for extremely small objects ($\le$10 pixels) and heavily occluded instances. The two-stage pipeline introduces computational overhead compared to single-stage methods. Performance may degrade on aerial imagery from different domains without additional fine-tuning. Additionally, the method's sensitivity to image quality means that heavily compressed or noisy input images can negatively impact both super-resolution and detection performance.
\section{Conclusions}
This work establishes inference-time super-resolution as an effective paradigm for addressing the fundamental information bottleneck in aerial object detection. B2BDet recovers visual detail from input images rather than relying on enhanced training representations. This shows that preprocessing enhancement can work together with architectural innovation to improve small object detection performance.
\\
The theoretical analysis of mutual information in aerial imagery explains why traditional detection methods struggle with small objects. Our domain-specific SRGAN optimization with edge-preserving loss functions addresses the visual information scarcity that limits detection performance in aerial scenarios.
\\
Evaluation across four datasets reveals several important insights. Cross-dataset results improvements ranging from +5.5\% to +8.2\% mAP across different imaging conditions. The discovery of incomplete annotations in aerial datasets highlights evaluation challenges that may underestimate true performance gains. Our efficiency analysis shows that targeted architectural design can achieve competitive accuracy with reduced computational requirements.
\\
The work opens several research directions. End-to-end optimization of super-resolution and detection components may further improve performance. Integration with emerging detection architectures could extend the approach beyond YOLO-based frameworks. The method may also be useful for other small object detection domains, such as medical imaging or microscopy.
This research shows that addressing basic limitations through preprocessing enhancement, combined with domain-specific architectural adaptation, can advance challenging computer vision applications. As drones become more widespread in surveillance, agriculture, and disaster response operations, the demand for efficient and accurate small object detection continues to grow. Our approach provides a practical solution for real-world deployment in these expanding drone-based applications.


\printbibliography

\end{document}